\documentclass[runningheads]{llncs}
\usepackage{cite}
\usepackage{amsmath,amssymb,amsfonts}
\usepackage{esvect}
\usepackage{algorithmic}
\usepackage{graphicx}
\usepackage{textcomp}
\usepackage{xcolor}
\usepackage{booktabs}
\usepackage{caption}
\usepackage{subcaption}
\usepackage{multirow}
\usepackage{pgfplots}
\usepackage{pgfplots}
\usepackage{multirow}
\usepackage{tabularx}
\usepackage{tikz}
\usepackage{tcolorbox}
\usepackage{fancyhdr}
\usepackage{float}
\usepackage{placeins} % 用于 \FloatBarrier
\usepackage{subcaption} % For subfigures
\usepackage{geometry}
\usepackage{authblk}
\pgfplotsset{compat=1.18}
\begin{document}
\title{YOLO-pdd: A Novel Multi-scale PCB Defect Detection Method Using Deep Representations with Sequential Images}
\titlerunning{Abbreviated paper title}
\author{Bowen Liu\inst{1}\textsuperscript{*}\thanks{{*}These authors contributed equally to this work.}%\orcidID{0000-1111-2222-3333}%
\and
Dongjie Chen\inst{2}\textsuperscript{*}\thanks{{*}These authors contributed equally to this work.}%\orcidID{1111-2222-3333-4444}%
\and
Xiao Qi\inst{3}\textsuperscript{\dag}\thanks{{\dag}\email{(Corresponding author)}}%\orcidID{2222--3333-4444-5555}%
}
\authorrunning{Bowen Liu et al.}
\institute{University of Southampton, Southampton, United Kingdom \and
University of Chinese Academy of Sciences, Beijing, China \and
%\url{http://www.springer.com/gp/computer-science/lncs} 
China Electronics Standardization Institute, Beijing, China\\
}

\maketitle  
% Typeset the header of the contribution

\begin{abstract}
With the rapid growth of the PCB manufacturing industry, there is an increasing demand for computer vision inspection to detect defects during production. Improving the accuracy and generalization of PCB defect detection models remains a significant challenge. This paper proposes a high-precision, robust, and real-time end-to-end method for PCB defect detection based on deep Convolutional Neural Networks (CNN). Traditional methods often suffer from low accuracy and limited applicability. We propose a novel approach combining YOLOv5 and multiscale modules for hierarchical residual-like connections. In PCB defect detection, noise can confuse the background and small targets. The YOLOv5 model provides a strong foundation with its real-time processing and accurate object detection capabilities. The multi-scale module extends traditional approaches by incorporating hierarchical residual-like connections within a single block, enabling multiscale feature extraction. This plug-and-play module significantly enhances performance by extracting features at multiple scales and levels, which are useful for identifying defects of varying sizes and complexities. Our multi-scale architecture integrates feature extraction, defect localization, and classification into a unified network. Experiments on a large-scale PCB dataset demonstrate significant improvements in precision, recall, and F1-score compared to existing methods. This work advances computer vision inspection for PCB defect detection, providing a reliable solution for high-precision, robust, real-time, and domain-adaptive defect detection in the PCB manufacturing industry.

\end{abstract}
\section{Introduction}
In modern electronic components, printed circuit boards (PCBs) represent a unique category characterized by trends toward intelligence, lightweight design, high density, and highly inclined equipment \cite{b1}. Given that PCB defects can compromise the functionality and safety of electronic devices, leading to reduced corporate yields, ensuring the quality and reliability of PCBs is crucial. This guarantee helps maintain the performance and longevity of electronic products \cite{b3}.

Despite advances in automatic optical inspection (AOI) and the emergence of deep learning-based detectors for PCB defect detection, several research gaps and challenges persist within this domain: 

\newcommand{\myitem}[1]{\item[(#1)]}
\begin{enumerate}
  \myitem{1} Robustness to Manufacturing Variability: Existing methods, including AOI and traditional deep learning detectors, often struggle to maintain high detection accuracy in the face of manufacturing variability, such as variations in PCB layout, component placement, and soldering conditions. Robust defect detection techniques are needed that can handle these variations effectively. 
  \myitem{2} Domain Adaptation and Generalization: Although deep learning-based detectors show promise in detecting defects, they can struggle with domain adaptation when deployed in diverse manufacturing environments or when confronted with new types of defects not seen during training. For example, DeepPCB \cite{b37} uses black-and-white images, whereas HRIPCB \cite{b38} employs color image representation. This difference in imaging methods and image representation impacts the compatibility and usability of the datasets, thereby adversely affecting the training and performance of detection models.  Therefore, methods to enhance the generalizability of these models in different PCB configurations and manufacturing conditions are essential. 
  \myitem{3} Real-time performance\cite{b1}: Real-time defect detection remains a challenge, particularly for complex PCBs with densely packed components. Methods that balance accuracy with real-time processing speed are necessary for practical industrial applications. 
  \myitem{4} Anomaly Detection and Sample Scarcity: Traditional defect detection methods often rely on supervised learning with labeled datasets. However, labeled samples require significant resources. In the absence of preexisting labels, how to leverage unsupervised learning\cite{b36}, semisupervised learning \cite{b39}, zero-shot learning \cite{b40}, and transfer learning methods\cite{b41} to identify novel defects or anomalies and reduce the dependence on samples remains an unresolved issue. Therefore, developing methods for anomaly detection using unsupervised learning, semi-supervised learning, etc. techniques are a promising area for future research. Fortunately, data-augmentation algorithms can be utilized to augment samples and reduce labeling costs. For example, virtual data sets can be created based on Generative Adversarial Networks (GANs)\cite{b42} or using model-design software\cite{b43}. 
  \myitem{5} Integration of Multimodal Data: Current defect detection methods focus mainly on visual inspection. Integrating other modalities such as thermal imaging or X-ray data \cite{b46}could provide complementary information for more accurate defect detection, especially in complex PCB designs. 

\end{enumerate}

Addressing these research gaps will contribute significantly to the advancement of the state-of-the-art in PCB defect detection, ultimately improving product quality and reliability in the electronics manufacturing industry.
In the rapid development of electronic technology, there is a growing demand for efficient and accurate methods to detect PCB defects. Traditional inspection methods, such as manual visual inspection, often exhibit lower detection accuracy, higher costs, and lower efficiency when faced with heavy inspection tasks and complex inspection environments. Meanwhile, automated optical inspection (AOI) is significantly influenced by optical conditions and tends to have a higher false detection rate \cite{b2}. Therefore, there is an urgent need for innovative approaches capable of reliably identifying defects in various types and configurations of PCBs. 

Given the aforementioned challenges, this paper aims to address the demands of the PCB manufacturing industry for high precision, robustness, real-time performance, and domain adaptation by proposing a PCB defect detection method based on deep convolutional neural networks (CNNs). This method aims to overcome the deficiencies of existing PCB defect detection methods and improve the accuracy and generalization performance of the model. To address this issue, the research team introduces a novel approach that uses deep CNNs and integrates feature extraction, defect localization, and classification into a multistage network architecture. Extensive experiments on a large-scale PCB dataset demonstrate significant improvements in this method compared to existing approaches. Particularly noteworthy is its achievement of real-time performance, making it suitable for widespread industrial applications. Additionally, the study addresses the challenge of domain adaptation in PCB defect detection by addressing domain drift through feature alignment from data sources, and experimental validation confirms the method's robustness and adaptability across various manufacturing environments and conditions. Consequently, this research represents a significant breakthrough in computer vision technology for PCB defect detection, providing the manufacturing industry with a reliable solution for defect detection.

The contributions of this paper are significant and can be summarized as follows:

\begin{itemize}
    \item [\textcolor{black}{$\bullet$}] A novel framework using a deep CNN-based, multi-stage network for PCB defect detection has been proposed, integrating feature extraction, defect localization, and classification into a unified system.
    \item [\textcolor{black}{$\bullet$}] A high-performance CNN-based method that leverages both global and local features to significantly improve PCB defect detection has been developed, achieving notable gains in precision, recall, and F1-score.
    \item [\textcolor{black}{$\bullet$}]A temporal approach with domain adaptation for PCB defect detection has been proposed, integrating YOLOv5 and Res2Net modules to ensure robustness under blurred or occluded conditions, significantly outperforming baseline methods.

\end{itemize}

\section{Related works}
In modern industrial scenarios, as the core component of electronic equipment, the printing circuit board (PCB) plays a vital role in ensuring the reliability and stability of the product\cite{b44}. However, in the entire manufacturing and application of PCB, various defects and problems are often encountered. In order to ensure the quality of PCB, the industry uses various methods to perform defect detection and electricity testing. 
\subsection{Manual Defect Detection Methods}
Defect detection has been a crucial aspect in the field of PCB manufacturing. Traditional manual visual inspection stands as one of the primary methods employed. This method relies heavily on operators that use tools such as magnifying glasses or calibrated microscopes to inspect the quality of circuit boards, making judgments based on their skills and experience with respect to the need for rectification \cite{b20}. However, this visually dependent inspection method exhibits certain limitations, with subjective estimation being the most noticeable risk. Subjective judgments of the operator can lead to erroneous detection results, affecting the quality and performance of the product, increasing the risk of industrial accidents such as fires caused by cable shorts or connection errors, and consequently reducing the safety of the production line \cite{b8}. Moreover, this method highly depends on the skills and experience of operators, necessitating factories to allocate substantial financial resources to recruit proficient personnel. Even the most experienced experts may err due to subjective estimation, resulting in inaccurate detection outcomes \cite{b21}. Therefore, manual visual inspection methods are increasingly becoming unsuitable for practical production tasks \cite{b19}.

\subsection{Automatic Optical Inspection} 
AOI based on machine vision (AOI) is a technology utilized for the detection of surface defects and abnormalities present on electronic components. It employs cameras and image processing algorithms to scan and analyze components and connection points on the printed circuit board (PCB), aiming to identify potential defects such as short circuits, open circuits, poor soldering, incorrect soldering, etc. \cite{b24}.
Specifically, the operational principle of AOI (Automatic Optical Inspection) is rooted in machine vision technology, employing red, green, and blue LED lights to illuminate PCB components \cite{b22}. It utilizes the principle of optical reflection to depict the soldering conditions of components on the PCB. Subsequently, it employs cameras or sensors to capture and analyze images of the printed circuit board (PCB) for the detection of defects within PCB components. Additionally, this method incorporates various innovative and improved techniques, such as automatic X-ray inspection (AXI), two-dimensional automated optical inspection (2D AOI), and three-dimensional automated optical inspection (3D AOI), for PCB defect recognition \cite{b1}. 
However, current automatic optical inspection (AOI) techniques are susceptible to the effects of ambient light, leading to significant distortion in captured images and an increase in misjudgment rates during the detection process \cite{b23}. In order to address this issue, although many PCB manufacturers opt for manual reevaluation, this undoubtedly incurs additional labor costs, resulting in resource wastage.

\subsubsection{Machine Learning-based Detectors} 
To overcome the time-consuming nature and reliance on a large number of reference images associated with traditional image processing methods, many researchers have turned to machine learning algorithms \cite{b33}. The current main research methods for PCB image defect detection based on machine learning can be divided into supervised learning (BP neural network, SVM), unsupervised learning (PCA, clustering methods), and semi-supervised learning \cite{b25} \cite{b26} \cite{b27}. These algorithms have become a focal point of research in the field of PCB defect detection, encompassing numerous classical machine learning algorithms. Through these methods, the detection of solder joints and components can be performed more rapidly and accurately, offering new possibilities for quality control in the PCB manufacturing process. Furthermore, despite the capability of image subtraction-based methods to achieve high-precision detection of PCB defects, machine learning algorithms excel in enhancing efficiency and accuracy, making them a viable choice in certain circumstances \cite{b4}.
A common approach, as explained by Yun et al. \cite{b9}, utilizes a Support Vector Machine (SVM) in conjunction with a tiered circular illumination technique to inspect PCB solder joints. The researchers assembled a dataset comprising 402 solder joints for training and testing the model, achieving defect detection rates of 96. 07\% and 98\%, respectively. In particular, this performance exceeds that of alternative methods, including the K-means classifier \cite{b10}  and the back-propagation (BP) classifier \cite{b11}. However, it is imperative to acknowledge that this method encounters several challenges, such as the selection of hyperparameters C and kernel in SVM, alongside the necessity for a sophisticated three-color circular illumination system.

\subsubsection{Deep Learning-based Detectors} 
Deep learning methods, especially convolutional neural networks (CNNs) including deep CNN (DCNN)\cite{b12},  VGG neural network \cite{b32}, \cite{b31} and ResNet neural network\cite{b11}, \cite{b34}, etc., have been widely applied in various fields such as image processing, classification, object detection, and segmentation\cite{b28}. These methods automatically extract image features, simplifying the image preprocessing process and effectively enhancing the accuracy and speed of detection. Compared to traditional detection methods, CNN-based approaches demonstrate strong robustness to environmental factors and noise. Across diverse datasets, CNN-based object detection algorithms exhibit excellent performance, serving as the primary driving force in the advancement of object detection. Consequently, deep learning algorithms have garnered the attention of researchers for PCB defect detection, achieving commendable detection performance \cite{b4}. 

In general, PCB detection methods based on deep learning can be classified into two categories: two-stage and single-stage approaches. Among the most commonly used methods for detecting PCB defects are the two-stage target detection algorithms: Region-Based Convolutional Network (R-CNN) \cite{b48} and one-stage target detection algorithms: You Only Look Once (YOLO) series\cite{b49}, Single shot multi-box detector (SSD) \cite{b50}. The important point to note is that Faster R-CNN \cite{b5} is an improved object detection model developed based on R-CNN \cite{b6} and Fast R-CNN \cite{b7}. Compared to the two-stage approach, the YOLO series, although relatively less precise, boasts a higher detection speed. Furthermore, due to its end-to-end architecture, the YOLO series is relatively simple to implement and is thus widely favored in the industrial domain. 
Ding et al. \cite{b12} introduced TDD-net, a novel micro-defect detection network based on faster R-CNN. Inspired by YOLOv2 \cite{b13}, the method autonomously determines anchor scales suitable for PCB images. By employing various data augmentation techniques and the Feature Pyramid Network (FPN) \cite{b14}, high detection accuracy (98.9\% mAP) is achieved. However, TDD-net operates as a two-stage framework, resulting in a large parameter count and a slower detection speed. The use of fully connected layers increases the complexity of the model, while RoI pooling may compromise the translational invariance of features. Despite its strong performance in defect detection, challenges remain in assessing real-time applicability and sensitivity to parameters.
CFIL \cite{b9} proposes a frequency-domain feature extraction module and feature interaction in the frequency domain to enhance salient features.
MFC \cite{b11} proposes a frequency-domain filtering module to achieve dense target feature enhancement.
Adibhatla et al.\cite{b17} utilized YOLOv5 \cite{b18} for PCB defect detection, improving accuracy through data augmentation and improved model backbone, with the largest model achieving a detection accuracy of 99.74\%. However, the method only distinguished between defective and non-defective targets without classifying defect types and exhibited low defect confidence scores. Additionally, to reduce computational costs, detection was performed only on small cropped images, potentially limiting real-time detection across entire PCB images.

\section{Methodology}
The primary objective of this research is to develop an advanced PCB defect detection method using deep convolutional neural networks (CNNs) to address the limitations of traditional inspection techniques. Specifically, our research aims to achieve the following objectives: (1) High Precision and Robustness: Develop a defect detection system that can achieve high precision and robustness, ensuring reliable identification of various types of PCB defects with minimal false alarms. (2) Real-time performance: Design the system to operate in real time, enabling rapid inspection and detection of defects during PCB manufacturing processes. (3) Domain Generalization and Efficiency: Investigate and improve domain adaptation techniques for PCB defect detection to enhance the efficiency and generalization of the model in diverse manufacturing environments and conditions. This comprehensive methodology aims to develop an effective and efficient PCB defect detection system using state-of-the-art deep learning techniques, tailored for deployment in real-world manufacturing environments.

\subsection{The proposed PCB defect detection framework}

The proposed PCB defect detection framework leverages deep convolutional neural networks (CNNs) to achieve high precision, robustness, real-time performance, domain adaptation, generalization, and efficiency. The framework consists of the following key components: input data handling, which includes pre-processing of PCB images to enhance quality, address noise, and standardize input dimensions; CNN architecture design, involving the development of a custom CNN architecture for PCB defect detection, comprising feature extraction, defect localization, and classification layers; integration of domain adaptation techniques to enable the model to generalize across different manufacturing environments and conditions; and optimization for real-time processing to ensure real-time performance during PCB inspection processes. Data collection and pre-processing techniques for PCB images involve the acquisition of a diverse dataset of PCB images representing various types of defects and manufacturing conditions. Since most PCB defect visual inspection systems, particularly those based on machine learning and deep learning techniques, are fundamentally data driven, addressing PCB defect data preprocessing constitutes a crucial step in the process. Specifically, preprocessing techniques include image augmentation (generation of synthetic data to increase dataset diversity and robustness), normalization and standardization (rescaling and normalizing pixel values to facilitate model training), and noise reduction (application of filtering and denoising techniques to enhance image clarity).

\subsection{Deep Representation Methods for PCB Defect Detection}

Our work focuses on capturing discriminative information from PCB images using advanced feature extraction methods. We leverage pre-trained CNNs, such as ResNet and VGG, for automatic feature extraction from PCB images. We employ localization techniques, using region-based or sliding-window approaches, to pinpoint defects within PCB images.

Our training strategies emphasize optimizing model performance and generalization. We select appropriate loss functions, such as categorical cross-entropy, to train CNNs on imbalanced defect datasets. To avoid overfitting, we apply regularization techniques such as dropout, batch normalization, and weight regularization. Additionally, we iteratively tuned model hyperparameters, including learning rate and batch size, to optimize performance metrics.

Understanding the importance of the loss function in machine learning and statistical modeling, we recognize it as a mathematical function that maps the difference between the predicted output of our model and the true labels to a non-negative real number, representing the prediction error or risk level of the model. Our goal is to minimize this loss function to optimize our model, enabling it to make more accurate predictions on unseen data.

For the YOLOv5 series, the loss function that we use consists mainly of four components: localization error, confidence error with object, confidence error without object, and classification error. By addressing these components, we enhance the accuracy and reliability of our model in PCB defect detection. By comprehensively considering the optimization of these four parts, we obtain the loss function of YOLOv5, which can be expressed as follows: 

\begin{equation}
    \begin{split}
L_{\text{YOLOv5}} = & \lambda_{\text{coord}} \sum_{i=0}^{S^2} \sum_{j=0}^{B} \mathbb{1}_{ij}^{obj}[(x_i - \hat{x}_i)^2 + (y_i - \hat{y}_i)^2] \\
& + \lambda_{\text{coord}} \sum_{i=0}^{S^2} \sum_{j=0}^{B} \mathbb{1}_{ij}^{obj}[(\sqrt{w_i} - \sqrt{\hat{w}_i})^2 \\
&+ (\sqrt{h_i} - \sqrt{\hat{h}_i})^2] \\
& + \sum_{i=0}^{S^2} \sum_{j=0}^{B} \mathbb{1}_{ij}^{obj}(C_i - \hat{C}_i)^2 \\
& + \lambda_{\text{noobj}} \sum_{i=0}^{S^2} \sum_{j=0}^{B} \mathbb{1}_{ij}^{noobj}(C_i - \hat{C}_i)^2 \\
& + \sum_{i=0}^{S^2} \mathbb{1}_{i}^{obj} \sum_{c \in \text{classes}} (p_i(c) - \hat{p}_i(c))^2\label{eq}
\end{split}
\end{equation}

During the training process of the YOLOv5 model, the back-propagation (BP) algorithm is employed. This widely used optimization algorithm calculates the gradients of the model parameters based on the loss function and updates the parameters using optimization methods such as gradient descent to minimize the loss function. 
In the process of this algorithm, comparing the actual output (y) with the expected output (r) yields an error signal. This error signal is propagated back through each layer of the neural network, propagating layer by layer, and computing the error signal for each layer. Subsequently, by adjusting the connection weights of each layer to minimize error, the performance of the neural network is optimized. 

\subsection{YOLOv5 combined with Res2Net module}

The Res2Net\cite{b45} architecture advances convolutional neural networks (CNN) by introducing a multiscale feature extraction mechanism. Integrating Res2Net with YOLOv5 enhances object detection by leveraging multiscale feature extraction, improving the detection of objects at different scales and resolutions\cite{b47}.
Fig.1 shows the architecture and workflow of a Res2Net Block.1. \textbf{Input Feature Splitting}: The input feature map is divided into several smaller segments.
2. \textbf{Hierarchical Transformation}: Segments are hierarchically transformed with increasing complexity.
3. \textbf{Feature Aggregation}: Transformed segments are concatenated, combining multi-scale features.
4. \textbf{Residual Connection}: A residual connection is added to stabilize the training and mitigate the vanishing gradient problem.

As Fig. 2 shows, our work provides an efficient Res2Net hierarchical approach, and multiscale transformations enhance CNNs' capabilities, making it a robust solution for complex PCB defect detection tasks. The following tips are some advantages of combining YOLOv5 and Res2Net for PCB defect detection:
1. \textbf{Enhanced Multi-Scale Feature Extraction}: Res2Net's ability to capture a wide range of features within a single block enhances the feature extraction capabilities of YOLOv5. This is particularly useful for detecting defects of various sizes in PCBs.
2. \textbf{Improved Detection Accuracy}: The combination improves the accuracy of defect detection, especially for small and intricate defects. Res2Net's multiscale processing complements the robust detection framework of YOLOv5.
3. \textbf{Real-time Processing}: YOLOv5 is known for its real-time object detection capabilities. When combined with Res2Net, the enhanced feature extraction does not significantly compromise processing speed, making the system suitable for real-time applications in industrial settings.
4. \textbf{Better Generalization}: The integration of advanced residual networks like Res2Net helps in better generalization across different types of PCBs and defects, improving the model's robustness and adaptability to various inspection scenarios.
5. \textbf{Scalability}: The combined framework is easily adjustable for specific tasks and computational constraints, allowing for scalable solutions that can be fine-tuned according to the requirements of different industrial applications.
6. \textbf{Efficient Training}: Using the strengths of both YOLOv5 and Res2Net can lead to more efficient training processes. Res2Net’s hierarchical residual-like connections can facilitate better gradient flow, potentially speeding up convergence during training.
7. \textbf{Robustness to Variability}: The combined approach is more robust to variability in PCB manufacturing processes. It can handle a wide range of defect types and environmental conditions, ensuring reliable performance in diverse operational environments.
8. \textbf{Comprehensive Detection Capabilities}: The synergy of YOLOv5’s object detection prowess with Res2Net’s feature extraction ensures complete defect detection, covering prominent and subtle defects that could be overlooked by other methods.

\begin{figure}[H]
\centering
\includegraphics[width=0.5\textwidth,scale=0.5]{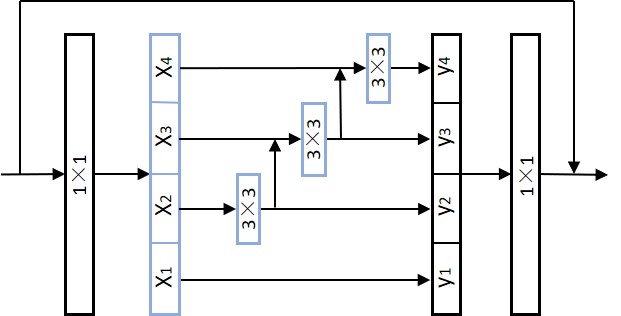}
    \caption{The framework of Res2net module}
    \label{figure.1}
\end{figure}

\begin{figure}[h]
    \centering
    \includegraphics[width=0.9\textwidth, scale=0.5]{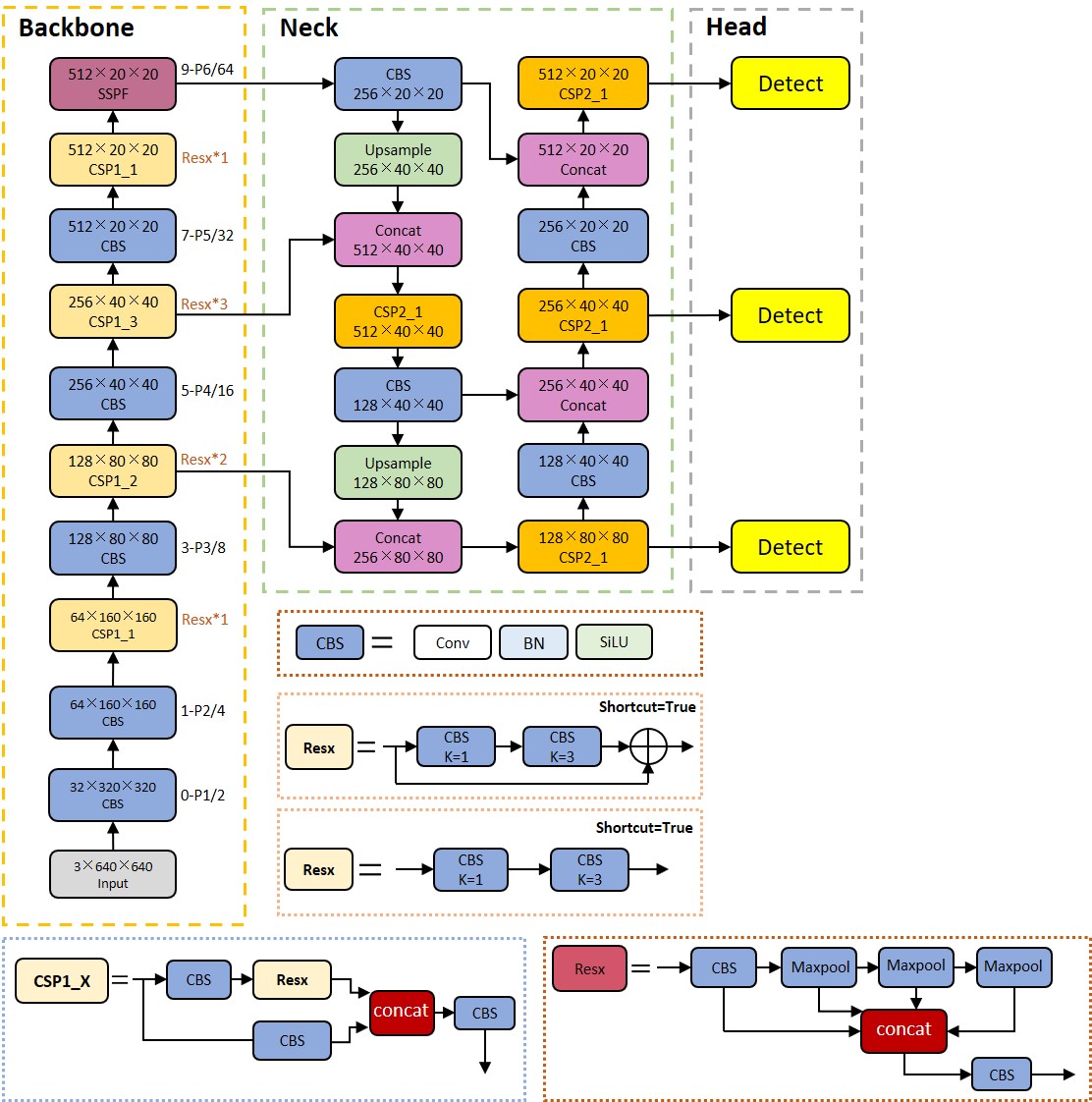}
    \caption{The main framework for YOLO-pdd defect detection}
    \label{figure.2}
\end{figure}

\subsection{Method Discussion}
The proposed method integrates the YOLO series with traditional PCB detection techniques, to combine the strengths of both approaches. The YOLO-pdd framework leverages YOLO's real-time processing capabilities and end-to-end architecture to achieve rapid defect detection without compromising accuracy. By incorporating the Res2Net architecture, our method significantly enhances the extraction of multiscale features, enabling the model to detect defects of varying sizes and complexities more effectively.

The implementation of a Feature Pyramid Network (FPN) with residual modules further boosts the model's ability to recognize PCB defects in sequential images from real industrial applications. The FPN allows the model to utilize features from different scales, ensuring that even small and subtle defects are detected. This multi-scale approach is particularly beneficial in PCB defect detection, where defects can range from minute cracks to larger structural issues.

Advanced data augmentation techniques are employed to increase the diversity and robustness of the training dataset. These techniques include generating synthetic data, rescaling, normalization, and noise reduction, all of which contribute to a more generalized and adaptable model. The use of optimal transport for domain generalization ensures that the model performs consistently across different manufacturing environments, addressing the variations in lighting, noise, and defect patterns that are common in real-world scenarios.

Optimization for real-time processing is another critical aspect of our method. By streamlining the model architecture and inference pipeline, we ensure that the YOLO-pdd framework can be deployed in high-speed production lines without causing delays. The high fps rate achieved by our model underscores its suitability for real-time applications, providing a significant advantage over traditional inspection methods that are often slower and less accurate.

Our experimental results on the PKU-Market-PCB dataset demonstrate the superiority of the YOLO-pdd framework in terms of accuracy, precision, recall, and processing speed. The significant improvements observed in these metrics highlight the effectiveness of our approach in improving defect detection performance. Furthermore, the robustness and adaptability of the YOLO-pdd model make it a valuable tool for various industrial applications, where consistent and reliable defect detection is crucial.

\section{Experimental results}
\subsection{Dataset}
The evaluation of our proposed PCB defect detection system is based on a comprehensive dataset composed of various PCB images obtained from real-world manufacturing processes. This dataset covers various types of defects commonly encountered in PCB, including short circuits, open circuits, poor soldering, and component misalignment. The data set is meticulously annotated with ground-truth labels for accurate model training and evaluation. 
This study used the publicly available PCB defect dataset (PKU-Market-PCB) released by the Open Lab on Human-Robot Interaction at Peking University. The original set of 1,386 images was expanded to 10,886 images by applying image augmentation techniques. To comprehensively assess the domain adaptation, generalization capabilities, and robustness, 6,000 images were then randomly sampled according to the six types of defects present in the data set (missing hole, mouse bite, open circuit, short, spur, and spurious copper). 

\subsection{Evaluation criteria}

To establish the efficacy of our proposed approach, we compare its performance with several baseline methods commonly used in PCB defect detection. These methods include traditional machine learning techniques, rule-based systems, and existing deep learning architectures. The baseline methods provide a benchmark for evaluating the effectiveness and efficiency of our proposed system. 

The comparison metrics involved are as follows:
\begin{itemize}
\item \textbf{The combinations of model classification categories} 
\\For binary classification tasks, combinations of true class labels and predicted class labels can be categorized into true positives (TP), true negatives (TN), false positives (FP), and false negatives(FN). Through these combinations, accuracy, recall, and the F1 score can be calculated. 
\item \textbf{Accuracy} 
\\Accuracy is defined as the ratio of true positives to the sum of true positives and false positives:
\begin{equation}
\text{Accuracy} = \frac{\text{TP}}{\text{TP} + \text{FP}}\label{eq}
\end{equation}

\item \textbf{Recall}
Recall is the proportion of true positive predictions out of all actual positive instances.
\begin{equation}
\text{Recall} = \frac{\text{TP}}{\text{TP} + \text{FN}}\label{eq}
\end{equation}

\item \textbf{F1-score}
F1-score is the harmonic mean of precision and recall, balancing the trade-off between precision and recall.
\begin{equation}
\text{F1-score} = 2 \times \frac{\text{Accuracy} \times \text{Recall}}{\text{Accuracy} + \text{Recall}}\label{eq}
\end{equation}

\end{itemize}

\subsection{Experimental Evaluation}

To validate the effectiveness of the proposed PCB defect detection method, we performed comprehensive experimental evaluations. These evaluations were performed on a workstation equipped with an Intel i7-9800K CPU, 64GB RAM, and a NVIDIA RTX2080Ti GPU. The model was trained using a PyTorch framework. The main hyperparameters were set as follows: a learning rate of 0.001, 300 training epochs, and a batch size of 32.

We compared our experimental results with those existing methods reported in the literature. This section analyzes various performance metrics including the Precision, Recall curve, Precision-Recall curve, and F1-score curve to provide a comprehensive evaluation of the model's effectiveness.
As shown in Fig.3, the predefined accuracy of the six categories of PCB defect detection shows an ideal trend. Fig.3 presents the Precision curve, which illustrates the proportion of true positive predictions among all positive predictions made by the model. A high precision value indicates that the model makes fewer false positive errors. The curve demonstrates that the model maintains consistently high precision in different defect categories, suggesting its effectiveness in correctly identifying actual defects without generating excessive false alarms.
Fig. 4 depicts the recall curve, which measures the proportion of true positive predictions out of all actual positive instances. High recall indicates that the model is successful in detecting a large number of actual defects. The recall curve shows a favorable trend, with the model achieving high recall rates across various defect categories. This indicates that the model is capable of detecting most defects present in the PCB images, ensuring comprehensive defect coverage.
Fig.5 displays the Precision-Recall curve, which combines both precision and recall into a single plot. This curve provides a more detailed understanding of the trade-off between precision and recall for different decision thresholds. The area under the Precision-Recall curve (AUC-PR) is a useful indicator of overall model performance. The curve shows that the model maintains a good balance between precision and recall, achieving high AUC-PR values, indicating robust performance in identifying defects with minimal false positives and false negatives.
Fig. 6 illustrates the F1 score curve, which is the harmonic mean of precision and recall. The F1 score provides a single metric that balances the trade-off between precision and recall, offering a comprehensive measure of the accuracy of the model. The F1-score curve indicates that the model consistently achieves high F1-scores across different defect categories, reflecting its ability to accurately detect and classify PCB defects with both high precision and high recall.

\begin{figure}[h]
    \centering
    \begin{minipage}[b]{0.45\textwidth}
        \centering
        \includegraphics[width=\textwidth]{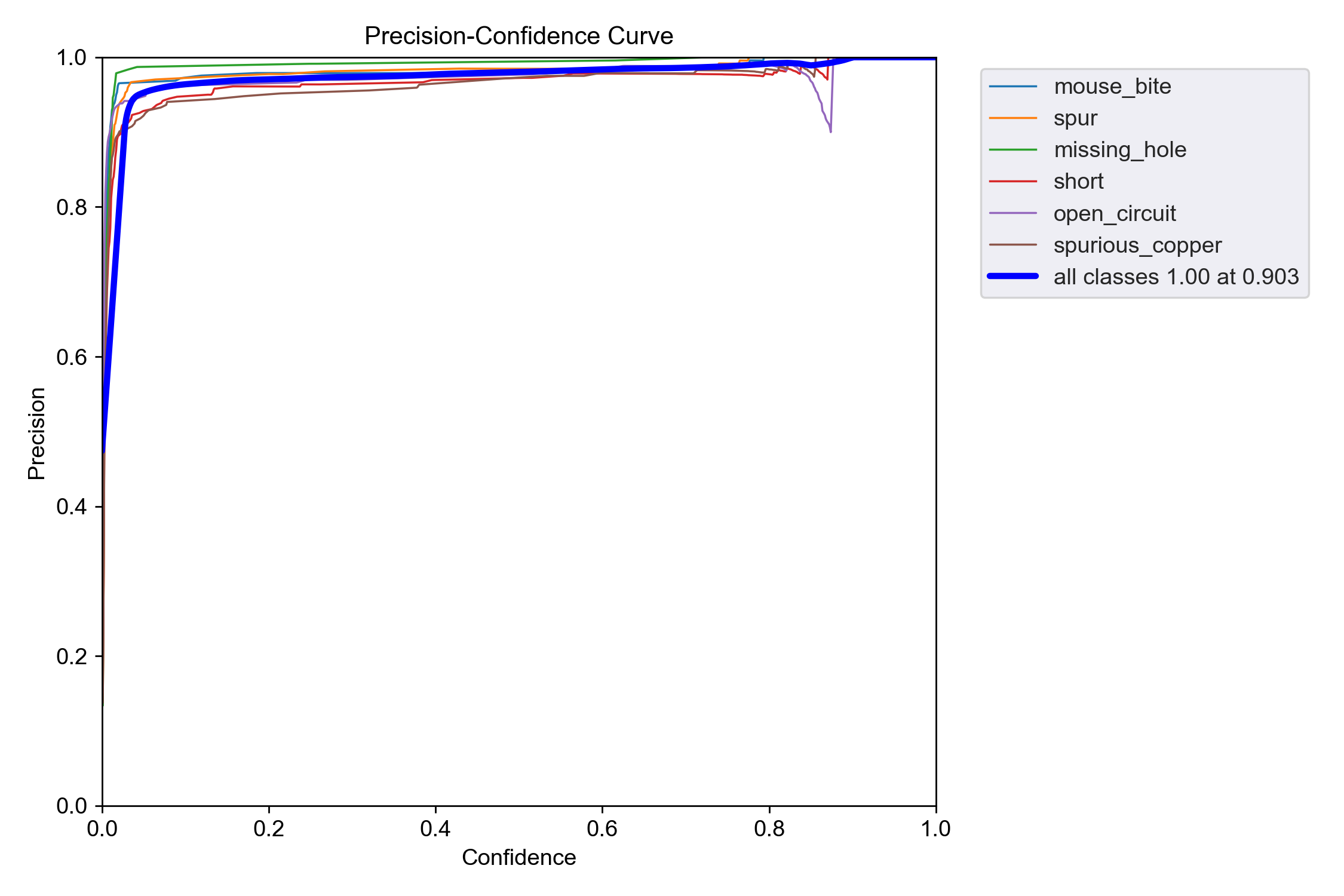}
        \caption{Precision curve}
        \label{figure.3}
    \end{minipage}
    \hspace{0.05\textwidth} % space between images
    \begin{minipage}[b]{0.45\textwidth}
        \centering
        \includegraphics[width=\textwidth]{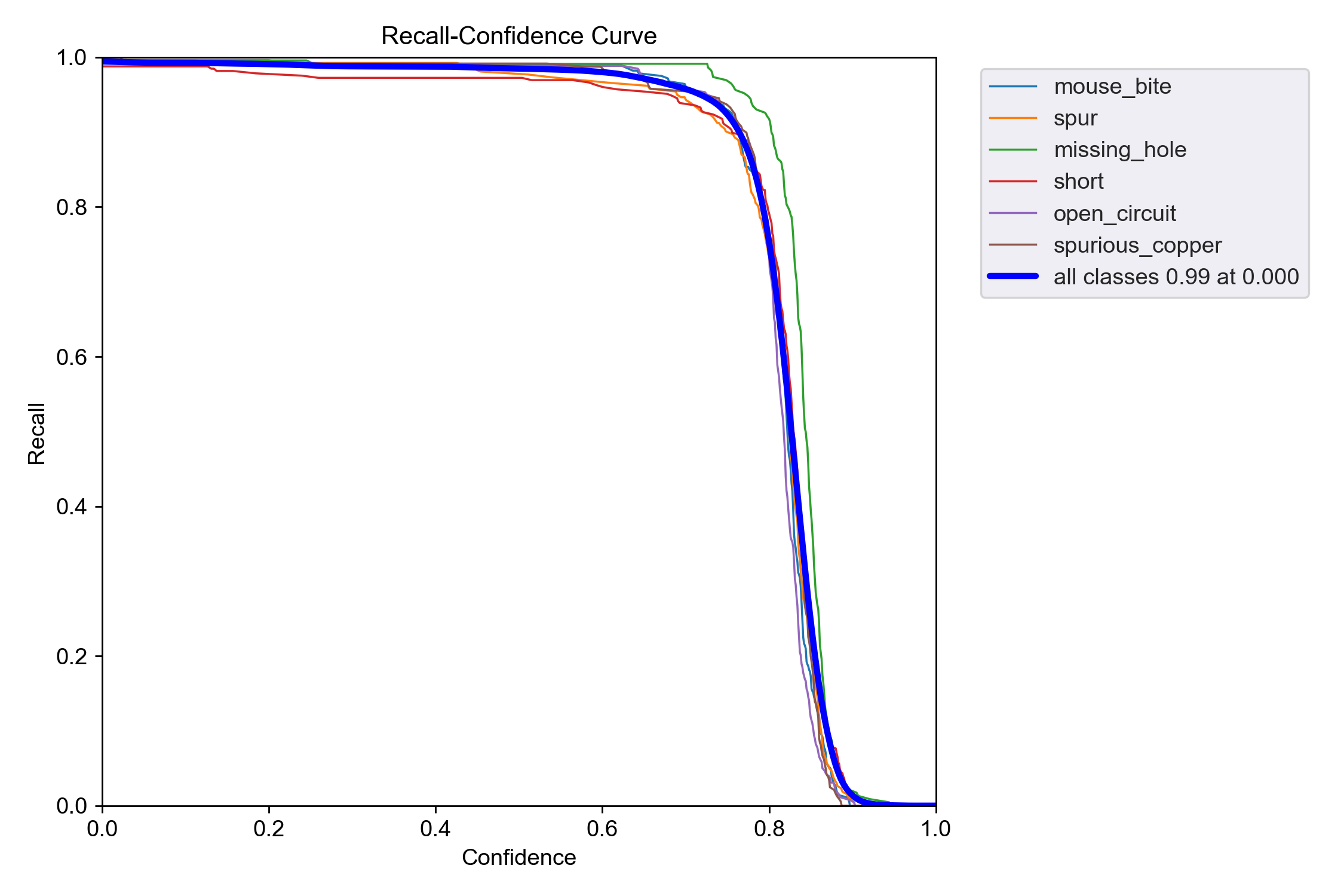}
        \caption{Recall curve}
        \label{figure.4}
    \end{minipage}
    \vspace{0.5cm} % space between rows
    \begin{minipage}[b]{0.45\textwidth}
        \centering
        \includegraphics[width=\textwidth]{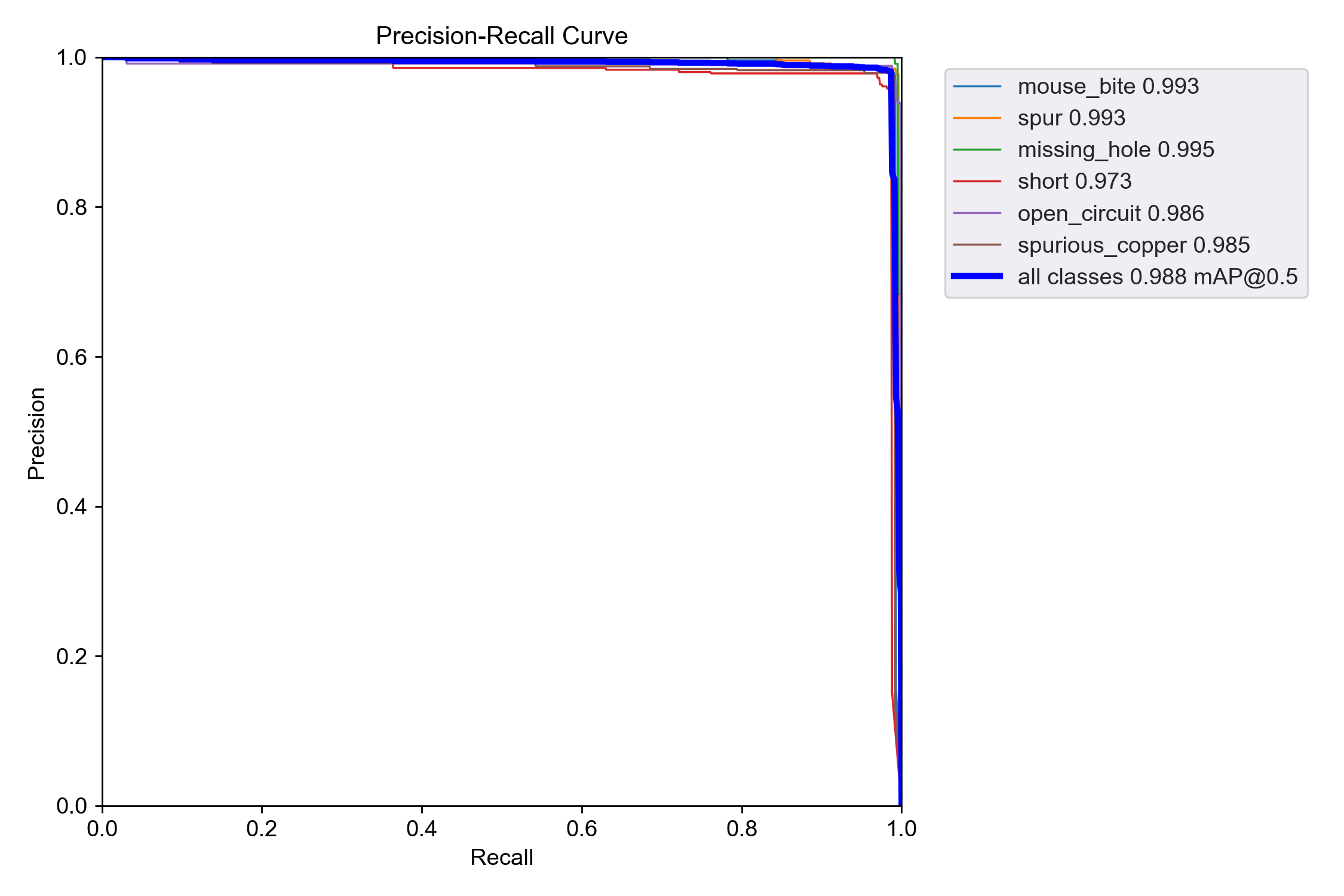}
        \caption{Precision-Recall curve}
        \label{figure.5}
    \end{minipage}
    \hspace{0.05\textwidth} % space between images
    \begin{minipage}[b]{0.45\textwidth}
        \centering
        \includegraphics[width=\textwidth]{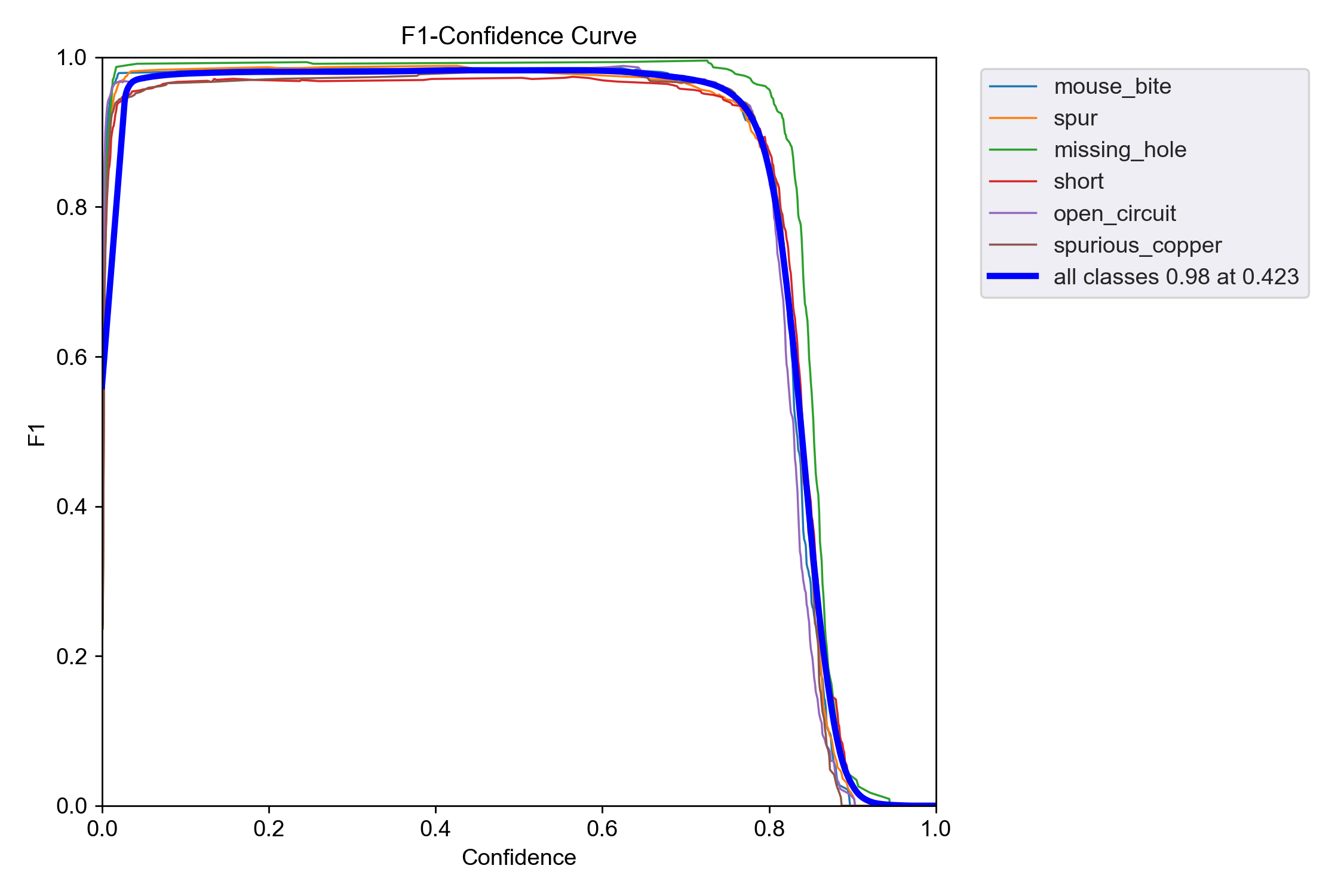}
        \caption{F1-score curve}
        \label{figure.6}
    \end{minipage}
\end{figure}

The indicators in Fig. 7 illustrate the outstanding performance of the YOLO-pdd framework in both recognition accuracy and processing speed. As shown in Fig. 7, YOLO-pdd significantly surpasses the original YOLO framework due to the superior multiscale feature extraction capabilities of the Res2Net architecture. The comparative analysis in Fig. 7 demonstrates that YOLO-pdd achieves a higher mean average precision (mAP) compared to other models, along with a higher frames per second (fps) rate, making it exceptionally suitable for real-time applications.

This notable improvement is attributed to the integration of Res2Net, which facilitates more detailed and robust feature representation, leading to improved defect detection performance. The innovative use of the Res2Net architecture, combined with effective data augmentation and domain adaptation strategies, ensures that YOLO-pdd can handle a wide range of defects in real-time scenarios. Importantly, among all models with an fps greater than 30, YOLO-pdd exhibits the best real-time processing performance. This ensures that the model not only detects defects with high accuracy but also keeps up with the fast pace of production lines, providing immediate feedback crucial for quality control and reducing waste.

From Table 1, it is evident that the YOLO-pdd model outperforms other methods across all metrics, particularly excelling in terms of recall and F1-score. The table presents a comparative analysis of the performance of different methods on the PKU-Market-PCB dataset, highlighting the superiority of the YOLO-pdd model. The experimental results obtained using the proposed approach demonstrate significant improvements in the model's defect detection accuracy, precision, recall, and frames per second (fps). Specifically, the YOLO-pdd model achieves higher accuracy, indicating its ability to correctly identify defects, and superior precision, reflecting fewer false positive detections. The high recall value suggests that the model effectively detects a large proportion of actual defects, and the elevated F1-score indicates a balanced performance between precision and recall. Furthermore, the high frame rate of the model demonstrates its efficiency and suitability for real-time PCB defect detection applications. These improvements underscore the robustness and effectiveness of the YOLO-pdd model in practical industrial scenarios.

Significant improvements in the performance metrics of the YOLO-pdd model can be attributed to its innovative integration of the YOLO architecture with Res2Net. This combination leverages YOLO's speed and end-to-end capabilities with the advanced feature extraction of Res2Net, enabling the model to handle a diverse range of PCB defects with high precision and accuracy. The incorporation of Res2Net facilitates multi-scale feature extraction, particularly beneficial for identifying defects of varying sizes and shapes commonly encountered in PCB inspections. This multi-scale approach ensures the model maintains high detection performance across different defect types, contributing to its robustness and adaptability in real-world scenarios.

Additionally, the YOLO-pdd model's ability to handle a wide range of defect types makes it a versatile tool for PCB inspection. Traditional methods often struggle with specific types of defect or require multiple models for complete coverage. In contrast, the YOLO-pdd model's unified approach simplifies the inspection process, providing a single, efficient solution for various defect detection needs. This versatility streamlines the inspection workflow and reduces the complexity and cost associated with maintaining multiple detection systems.

Another critical aspect of the YOLO-pdd model is its capability to be integrated into existing industrial systems with minimal modifications. The model's high frame rate ensures it can keep up with the rapid pace of production lines, thereby not becoming a bottleneck in the manufacturing process. This real-time detection capability is crucial to maintaining high productivity levels and ensuring immediate feedback for any defects detected, which is vital for quality control and waste reduction. Furthermore, the robustness and adaptability of the YOLO-pdd model make it suitable for various PCB designs and manufacturing environments, demonstrating its practical utility and effectiveness in real-world applications.

% Moreover, the YOLO-pdd model incorporates advanced data augmentation and domain generalization strategies, which further enhance its robustness. Using techniques such as optimal transport and domain adaptation, the model can be effectively generalized across different manufacturing environments, addressing challenges such as domain shift and variability in defect appearance. This ensures consistent performance even when the model is deployed in new or changing production settings.

The YOLO-pdd model also benefits from its ability to be trained on large-scale datasets, capturing a wide variety of defect types and scenarios. This extensive training enables the model to learn intricate patterns and subtle differences between type defects, which improves its detection capabilities. The combination of YOLO's real-time processing, Res2Net's multi-scale feature extraction, and advanced data augmentation techniques positions the YOLO-pdd model as a state-of-the-art solution for PCB defect detection, offering unparalleled accuracy, efficiency, and adaptability in industrial applications.

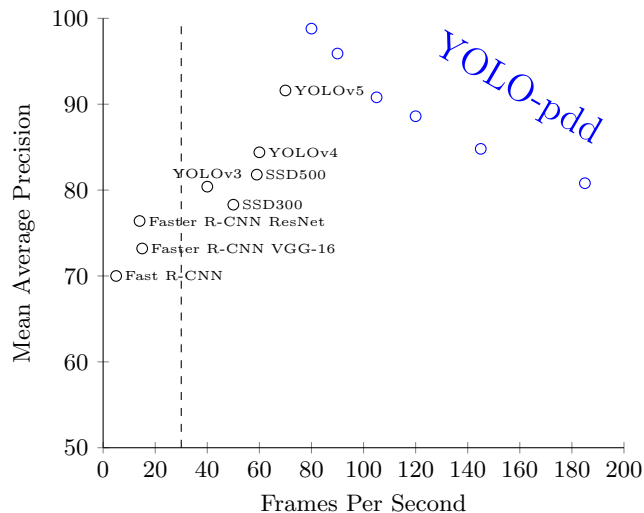
\begin{figure}[h]
    \centering
    \begin{tikzpicture}[scale=1] % 缩放整个图形，包括文本
        \begin{axis}[
            xlabel={Frames Per Second},
            ylabel={Mean Average Precision},
            xmin=0, xmax=200,
            ymin=50, ymax=100,
            xtick={0,20,40,60,80,100,120,140,160,180,200},
            ytick={50,60,70,80,90,100},
            legend pos=north west,
            major grid style={line width=.2pt,draw=gray!50},
            scatter/classes={
                a={mark=o,draw=black},
                b={mark=o,draw=blue}
            },
            axis lines=left, % 仅保留左边和下边的轴
            axis line style={-},
            tick align=outside,
            axis x line=bottom,
            axis y line=left,
            x axis line style={draw=black},
            y axis line style={draw=black},
            every outer y axis line/.append style={-}, % 移除右边框
            every outer x axis line/.append style={-}, % 移除上边框
        ]
        \addplot[scatter,only marks,scatter src=explicit symbolic]
        table[meta=label] {
            x y label
            5 70 a
            15 73.2 a
            14 76.4 a
            50 78.3 a
            59 81.8 a
            40 80.4 a
            60 84.4 a
            70 91.6 a
            185 80.8 b
            145 84.8 b
            120 88.6 b
            105 90.8 b
            90 95.9 b
            80 98.8 b
        };
        
        \node[anchor=west, font=\tiny] at (axis cs:70,91.6) {YOLOv5};
        \node[anchor=west, font=\tiny] at (axis cs:60,84.4) {YOLOv4};
        \node[anchor=west, font=\tiny] at (axis cs:5,70) {Fast R-CNN};
        \node[anchor=west, font=\tiny] at (axis cs:15,73.2) {Faster R-CNN VGG-16};
        \node[anchor=west, font=\tiny] at (axis cs:14,76.4) {Faster R-CNN ResNet};
        \node[anchor=west, font=\tiny] at (axis cs:50,78.3) {SSD300};
        \node[anchor=west, font=\tiny] at (axis cs:59,81.8) {SSD500};
        \node[anchor=south, font=\tiny] at (axis cs:40,80.4){YOLOv3} ;
        \node[anchor=south, font=\tiny] at (axis cs:67,72.9){};
        \node[anchor=south, font=\tiny] at (axis cs:140,99.8) {};
        \node[anchor=south, font=\tiny] at (axis cs:28,78.6){};
        \node[anchor=west, font=\tiny] at (axis cs:145,94.6){} ;
        \node[anchor=south west, font=\Large\color{blue},rotate=-30] at (axis cs:120,95) {YOLO-pdd};
        \draw[dashed] (axis cs:30,50) -- (axis cs:30,100);
        \end{axis}
    \end{tikzpicture}
    \caption{Accuracy and speed on PKU-Market-PCB dataset.}
    \label{fig:accuracy_speed}
\end{figure}

\vspace{-10pt} % 减少图和表之间的空间

\begin{table}[h]
    \centering
    \caption{PCB Defect Detection frameworks performance on PKU-Market-PCB dataset.}
    \setlength{\tabcolsep}{3pt} % Reduce the space between columns
    \resizebox{\linewidth}{!}{%
    \begin{tabular}{lccccccc}
    \toprule
    \multirow{2}{*}{\textbf{Model}} & \multirow{2}{*}{\textbf{Backbone}} & \multicolumn{6}{c}{\textbf{Evaluation Metrics}} \\
    \cmidrule(lr){3-8}
    & & \textbf{AP} & \textbf{AP50} & \textbf{AP75} & \textbf{Recall} & \textbf{F1-score} & \textbf{Fps}\\
    \midrule
    \multicolumn{8}{l}{\textbf{Two-stage methods}} \\
    \quad Faster R-CNN+++ & ResNet-101-C4 & 34.9 & 55.7 & 37.4 & 15.6 & 38.7 & 12 \\
    \quad Faster R-CNN w FPN & ResNet-101-FPN & 36.2 & 59.1 & 39.0 & 18.2 & 39.0 & 15 \\
    \quad Faster R-CNN by G-RMI & Inception-ResNet-v2 & 34.7 & 55.5 & 36.7 & 13.5 & 38.1 & 22 \\
    \quad Faster R-CNN w TDM & Inception-ResNet-v2-TDM & 36.8 & 57.7 & 39.2 & 16.2 & 39.8 & 23 \\
    \midrule
    \multicolumn{8}{l}{\textbf{One-stage methods}} \\
    \quad SSD513 & ResNet-101-SSD & 92.2 & 94.1 & 93.5 & 10.2 & 34.5 & 58 \\
    \quad DSSD513 & ResNet-101-DSSD & 93.6 & 95.3 & 94.2 & 13.0 & 35.4 & 66 \\
    \quad RetinaNet & ResNet-101-FPN & 91.1 & 94.1 & 92.6 & 21.8 & 42.7 & 50 \\
    \quad RetinaNet & ResNeXt-101-FPN & 91.8 & 95.1 & 93.4 & 24.1 & 44.2 & 51 \\
    \quad YOLOv4 & DarkNet-19 & 89.3 & 92.8 & 93.6 & 15.0 & 22.4 & 60 \\
    \quad YOLOv5 & Darknet-53 & 94.4 & 96.9 & 95.8 & 18.3 & 35.4 & 75 \\
    \quad \textbf{YOLO-pdd(ours)} & \textbf{Darknet-53+Res2Net} & \textbf{95.8} & \textbf{99.6} & \textbf{97.4} & \textbf{18.8} & \textbf{85.6} & \textbf{92} \\
    \bottomrule
    \end{tabular}%
    }
    \label{tab:comparison}
\end{table}

%\FloatBarrier % 确保图片不会影响前面的段落
\begin{figure}[H]
    \centering
    \includegraphics[width=\textwidth]{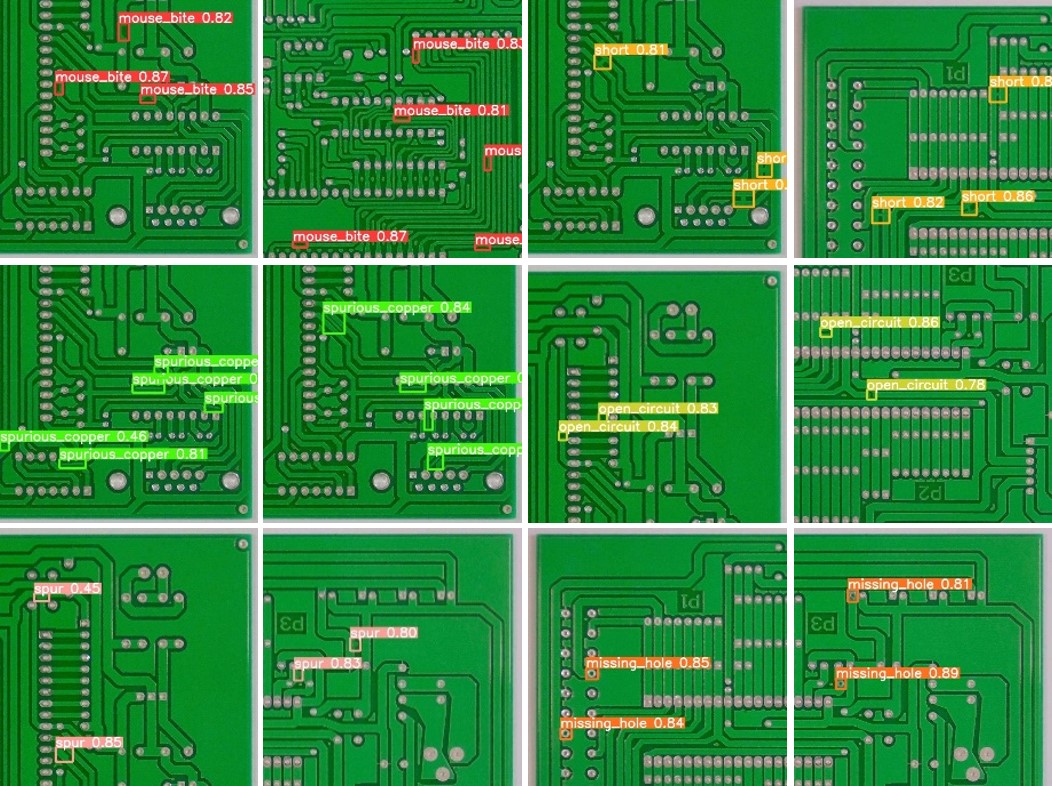}
    \caption{PCB Defect Detection results for sequential images}
    \label{fig:label1}
\end{figure}

\section{Conclusion}
In our paper, we introduce a novel framework that merges YOLOv5 and Res2Net for precise PCB defect detection in real industrial scenarios. By addressing the limitations of traditional methods and offering a high-performance, adaptable solution, the YOLO-pdd model sets a new standard for PCB defect detection, promising improved quality control and efficiency in the manufacturing industry. Our method not only achieves high accuracy and efficiency but also exhibits robustness and adaptability, making it well suited for practical industrial scenarios. Leveraging YOLO's real-time processing and end-to-end architecture ensures swift defect detection without sacrificing accuracy. Our method not only achieves high accuracy and efficiency but also demonstrates robustness and adaptability, offering potential solutions for reliable PCB inspection systems in various industrial contexts.

\end{document}